\documentclass[letterpaper, 10 pt, conference]{ieeeconf}  

\IEEEoverridecommandlockouts                              

\usepackage{xspace}
\usepackage{graphicx,grffile}
\usepackage{amsmath}

\usepackage{graphics} 
\usepackage{cite}

\title{\LARGE \bf
WAVE: Worm Gear-based Adaptive Variable Elasticity for Decoupling Actuators from External Forces
}

\author{Moses Gladson Selvamuthu$^{\dagger, 1}$, Tomoya Takahashi$^{2}$, Riichiro Tadakuma$^{1}$, and Kazutoshi Tanaka$^{2}$
\thanks{*This work was supported by JST, PRESTO Grant Number JPMJPR22C6, Japan.}
\thanks{
    $\dagger$Work done at OMRON SINIC X Corp. as part of an internship.
    $^{1}$Moses Gladson Selvamuthu and Riichiro Tadakuma are with Department of Mechanical Systems Engineering, Faculty of Engineering, Yamagata University, 4 Chome-3-16 Jonan, Yonezawa, Japan.
    {\tt\small mosesgladsonrocks@gmail.com}}%
    \thanks{$^{2}$Kazutoshi Tanaka and Tomoya Takahashi are with OMRON SINIC X Corporation, Hongo 5-24-5, Bunkyo-ku, Tokyo, Japan.
    {\tt\small kazutoshi.tanaka@sinicx.com}}%
}

\usepackage{fancyhdr}
\usepackage{url}

\begin{document}

\twocolumn[
\noindent
© 2025 IEEE. Personal use of this material is permitted. Permission from IEEE must be obtained for all other uses, in any current or future media, including reprinting/republishing this material for advertising or promotional purposes, creating new collective works, for resale or redistribution to servers or lists, or reuse of any copyrighted component of this work in other works.\\

\noindent
\textbf{Published article:}\\
M. Selvamuthu, T. Takahashi, R. Tadakuma, and K. Tanaka, ``WAVE: Worm Gear-based Adaptive Variable Elasticity for Decoupling Actuators from External Forces,'' 2025 IEEE/RSJ International Conference on Intelligent Robots and Systems (IROS 2025), 2025.
]

\thispagestyle{empty}
\pagenumbering{gobble}
\clearpage

\maketitle
\thispagestyle{empty}

\begin{abstract}

  Robotic manipulators capable of regulating both compliance and stiffness offer enhanced operational safety and versatility. Here, we introduce Worm Gear-based Adaptive Variable Elasticity (WAVE), a variable stiffness actuator (VSA) that integrates a non-backdrivable worm gear.
  By decoupling the driving motor from external forces using this gear, WAVE enables precise force transmission to the joint, while absorbing positional discrepancies through compliance.
  WAVE is protected from excessive loads by converting impact forces into elastic energy stored in a spring.
  In addition, the actuator achieves continuous joint stiffness modulation by changing the spring's precompression length.
  We demonstrate these capabilities, experimentally validate the proposed stiffness model, show that motor loads approach zero at rest--even under external loading--and present applications using a manipulator with WAVE.
  This outcome showcases the successful decoupling of external forces.
  The protective attributes of this actuator allow for extended operation in contact-intensive tasks, and for robust robotic applications in challenging environments.

\end{abstract}

\section{INTRODUCTION}

\begin{figure}
\centering
\includegraphics[width=0.92\linewidth]{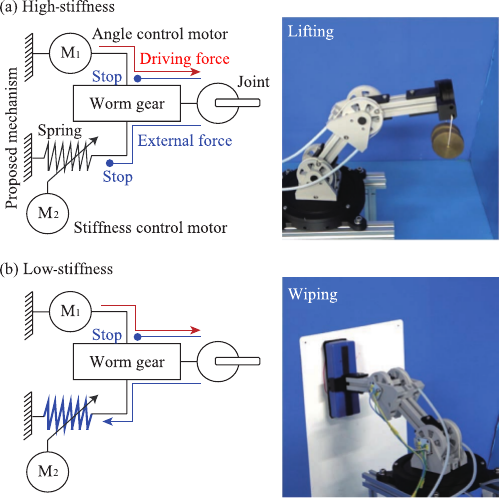}
\caption{The main idea of this study. Worm Gear-based Adaptive Variable Elasticity (WAVE) is a variable stiffness actuator for decoupling the angle control motor from external forces using a worm gear. (a) WAVE changes the joint stiffness between high-stiffness to support force and move precisely and (b) low-stiffness to compensate for position errors and absorb impact.}
\label{fig:teaser}
\end{figure}

Introducing robots into environments not specifically designed for them, such as human living spaces, requires flexibility to mitigate collision forces and prevent damage, while maintaining sufficient rigidity for precise and powerful actuation as shown in Fig.~\ref{fig:teaser} (right).
One effective approach is incorporating inherent compliance into the robot design, which helps absorb unexpected collision forces, enhancing safety and tolerance to positional errors.
However, in many scenarios, high payload capacity or accurate position control is also necessary for performing tasks at a level comparable to human capabilities.
Implementing variable joint compliance is beneficial in integrating these features into a single hardware system.

\begin{figure}
\centering
\includegraphics[width=\linewidth]{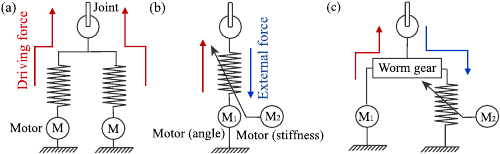}
\caption{Types of VSA principles. (a) Antagonistic. (b) Independent. (c) The proposed WAVE mechanism.}
\label{fig:vsms}
\end{figure}


A Variable Stiffness Actuator (VSA) is a hardware module designed to control compliance, enabling robots to change their joint stiffness in a mechanical manner to perform diverse tasks [1], [2].
Compared to software-based stiffness control, VSA has a fast response speed and can store impact force energy in the spring, protecting both the robot and the actuator from the force [3].
Based on actuator configuration as shown in Fig.~\ref{fig:vsms}, VSAs can be classified into the antagonistic type [4], [5] and independent type [6].
Antagonistic VSAs utilize two motors to control both elastic force and joint actuation through differential motion as shown in Fig.~\ref{fig:vsms}(a).
However, because the motors are constantly subjected to spring compression forces, additional torque is required beyond what is needed for joint movement. In contrast, as shown in Fig.~\ref{fig:vsms}(b), independent VSAs separate the roles of the angle control motor that primarily controls joint actuation and the stiffness control motor that adjusts spring precompression.
Since the VSA motor does not necessarily need to follow the angle control motor's speed during joint actuation, it can be smaller, reducing the overall mass of the two motors. Nevertheless, in terms of impact absorption, external impact forces transmitted through the spring can still reach the angle control motor, potentially leading to actuator damage. This study aims to develop a VSA that ensures the independence of force transmission paths for external impact absorption and joint actuation. Worm gears or power screws are known for their non-backdrivable properties and have been utilized in applications where large forces are applied [7].
A worm gear is a transmission component that converts rotational motion between perpendicular shafts, typically used to achieve high reduction ratios. A worm gear with small lead angle and single-start thread is non-backdrivable, where the worm wheel cannot drive the worm screw because the friction at the contact surface exceeds the torque generated by the reverse force.

In this paper, we propose a novel VSA mechanism named WAVE (Worm gear-based Adaptive Variable Elasticity) as shown in Fig.~\ref{fig:vsms}(c).
This mechanism incorporates a worm gear, which is typically fixed in the linear direction but, in our design, is constrained via a spring along that axis.
This allows for two distinct motion pathways: (A) rotation of the joint by the angle control motor ($\text{M}_{\text{1}}$), shown in red in Fig. 1(a), and (B) precompression of the linear spring by external force feedback or the stiffness control motor ($\text{M}_{\text{2}}$), shown in blue in Fig. 1(b).
As shown in Fig.~\ref{fig:teaser} (left), this mechanism transmits the angle control motor's torque to the joint via the worm gear while the spring passively absorbs displacement. This allows for adaptive trajectory generation in response to environmental interactions. Additionally, since the worm gear completely decouples the angle control motor, the impact force is immediately converted into elastic energy of the spring, effectively protecting the actuator. Furthermore, active compliance control--referring to the ability to adjust passive joint deflection in response to external forces by modifying the spring preload--enables the system to support high payloads.


In this paper, we introduce WAVE as a novel variable stiffness actuator that decouples the angle control motor from external forces through a worm gear mechanism.
The main contributions of our work are threefold:
\begin{enumerate}
    \item We propose a compact and integrated hardware design that leverages non-backdrivability to isolate the angle control motor from external forces.
    \item We demonstrate how active control of the worm gear's linear axis enables continuous switching between high-stiffness for precise and powerful actuation, and low-stiffness for safe interaction and impact absorption.
    \item Through experimental evaluation, we show that our mechanism can robustly handle a wide range of tasks, from high-load manipulation to contact-rich operations, without compromising safety or performance.
\end{enumerate}

\section{Related Work}

\subsection{Stiffness Control by Independent Motor}

To address the limitations of antagonistic VSAs, several designs using independent motor configurations have been proposed. Eiberger et al. introduced a quasi-antagonistic VSA, where the actuators for joint motion and spring length adjustment are placed separately [8]. Although the spring placement remains antagonistic, modifying the spring length shifts the equilibrium point, requiring coordinated control of both actuators. Vo et al. proposed a mechanism using a linkage system to adjust spring displacement, enabling complete independence between stiffness modulation and joint actuation [9].
Some designs implement distinct motion pathways for transmitting driving force and adjusting stiffness. Several studies have proposed modifying the pivot point of a lever mechanism to achieve variable stiffness by adjusting the moment arm [10], [11]. In these systems, the driving force is transmitted through lever rotation, while the pivot point is linearly adjusted by a stiffness control motor, thereby controlling the spring load on the joint. Sariyildiz et al. proposed a design where driving torque is transmitted via rotation, and stiffness is modulated by displacing a leaf spring parallel to the rotational axis [12], effectively varying the spring constant for more flexible stiffness control.

While these designs successfully decouple the stiffness control motor, they do not fully decouple the angle control motor.
Although the method of separating driving force and compliance into two directions is effective, using multiple components to achieve this can complicate the mechanism and increase its weight.
In contrast, the proposed mechanism achieves this separation using only the worm gear, simplifying the design and reducing the overall complexity.

\subsection{Non-Backdrivable VSA}
Mechanisms utilizing worm gears and other non-backdrivable components have been explored to prevent external force feedback to the motor. One approach uses an antagonistic worm gear mechanism that enables both joint rotation and spring compression through linear movement [13], [14], while another employs an overrunning clutch--a one-way mechanism that blocks reverse torque--to achieve similar non-backdrivable behavior. These mechanisms are primarily based on antagonistic configurations [15], and no independent designs have been reported that effectively separate driving torque from external forces. Among non-backdrivable mechanisms, worm gears also provide the benefit of integrated speed reduction while minimizing the number of components.

\subsection{VSA Module Integration and Comparison}
%

In cable-driven manipulators, various configurations exist for integrating VSAs, and the choice of configuration significantly influences the effectiveness of stiffness modulation. Several studies have investigated pulley-based mechanisms where springs are attached to pulleys, allowing easier integration than direct motor attachment [16], [17], [18]. However, under high loads, cable elongation due to constant spring preload can introduce modeling inaccuracies in joint angles and stiffness, making precise control more difficult. In soft robotics, structural VSAs are used to achieve high compliance and passive flexibility [19], [20], but these systems are limited by material constraints, reducing their suitability for high-load applications.

In this study, we adopt a motor-integrated modular approach that enables easy detachment and reassembly. By incorporating both softness and spring preloading within the module, it can be separated from the manipulator to improve adaptability. This modular configuration offers practical benefits across a range of robotic applications.

\section{WAVE: Worm gear-based Adaptive Variable Elasticity}

\begin{figure}
\centering
\includegraphics[width=0.8\linewidth]{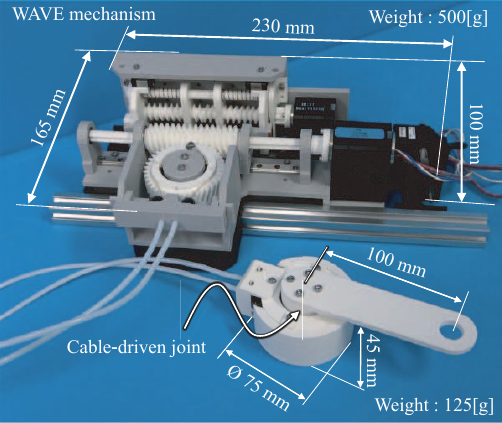}
\caption{Appearance, dimensions, and weights of WAVE and the joint actuated by WAVE via Bowden cables.}
\label{fig:appearance}
\end{figure}

In this section, we introduce WAVE.
Fig.~\ref{fig:appearance} shows the appearance, dimensions, and weights of WAVE and the joint actuated by WAVE via cables.
In this study, we leverage WAVE as an actuator for a cable-driven manipulator.

\subsection{Design}

\begin{figure}
\centering
\includegraphics[width=0.88\linewidth]{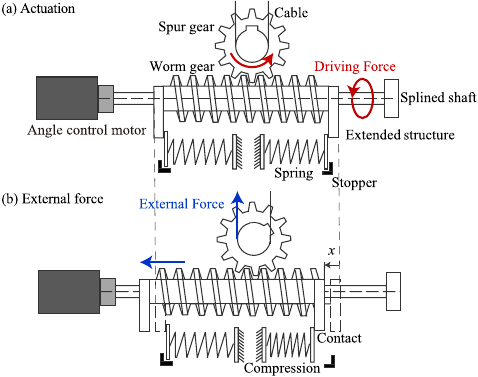}
\caption{Working principle of the WAVE. (a) Active joint rotation through worm rotation. (b) Passive sliding of worm gear under external force.}
\label{fig:compression}
\end{figure}

Fig.~\ref{fig:compression} illustrates the basic working principle of WAVE. The worm gear enables two degrees of freedom: rotation and linear axial displacement. A splined shaft with longitudinal slots meshes with matching features in the worm gear, allowing torque transmission while permitting axial motion. Fig.~\ref{fig:compression}(a) shows the condition when the shaft is rotated by the angle control motor, transmitting torque to the mating spur gear, which drives a pulley wound with cable. This rotation actuates the joint connected via Bowden cables. Fig.~\ref{fig:compression}(b) shows the condition when an external force is applied to the joint. The cable transmits this force back to the spur gear, causing axial sliding of the worm gear. Due to the non-backdrivability of the worm gear, the external force is not transmitted to the angle control motor, ensuring smooth decoupling of the angle control motor from external forces. Axial sliding is constrained by two springs positioned on either side of the worm gear, allowing resistance to both clockwise and counterclockwise external forces.

\begin{figure}
\centering
\includegraphics[width=0.88\linewidth]{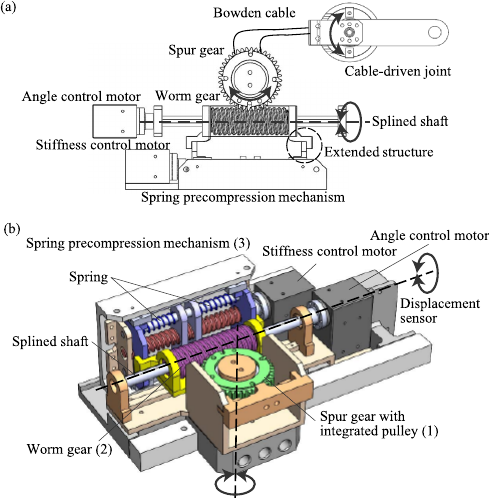}
\caption{Basic design of the VSA. (a) Basic mechanism of WAVE. (b) Schematics of WAVE.}
\label{fig:schematic}
\end{figure}

\begin{figure}
\centering
\includegraphics[width=0.88\linewidth]{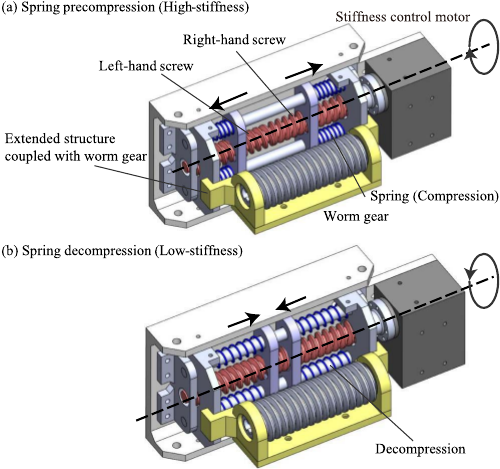}
\caption{Spring precompression mechanism. (a) Spring precompression. (b) Spring decompression.}
\label{fig:precompression}
\end{figure}

Fig.~\ref{fig:schematic}(a) shows the mechanism's basic design, which consists mainly of three parts: (1) a spur gear with an integrated cable pulley mating with the worm gear, (2) a 2-DOF worm gear assembly on a splined shaft, and (3) a spring precompression mechanism to control linear displacement of the worm gear.
Fig.~\ref{fig:schematic}(b) shows the schematic of the VSA. The linear displacement of the worm gear is coupled with the spring precompression mechanism through an extended structure. Fig.~\ref{fig:precompression} shows the precompression mechanism of WAVE.
Precompression of a spring means applying an initial force to shorten the spring from its free length, which ensures that the spring consistently applies a specific, constant force to the mechanical system.

When the worm gear moves to the right, the extended structure on the right side engages and compresses the springs positioned on that side, while the springs on the left remain unaffected and vice versa. The precompression length of the spring controls the rate of worm gear displacement, with greater precompression resulting in higher resistance and reduced displacement. The precompression length is adjusted by a separate stiffness control motor using a dual-hand screw arrangement. When the stiffness control motor rotates clockwise, one nut moves right and the other moves left and vice versa. This linear motion of the nuts is used to precompress or decompress the springs. Since two sets of springs are arranged in parallel, the equivalent spring stiffness is the sum of the individual spring stiffnesses.


The cable-driven joint with WAVE is shown in Fig.~\ref{fig:appearance}.
The worm gear was printed using Polylactic Acid (PLA), and the splined shaft for power transmission was fabricated from white resin using a FormLabs Form 3 printer. The worm gear is driven by a servo motor (Dynamixel XC330-T181-T, ROBOTIS), with the worm gear providing an additional 40:1 reduction that significantly increases the output torque. The components of the precompression mechanism are also 3D-printed from PLA and driven by a separate servo motor (Dynamixel XH540-W150-R, ROBOTIS). The worm and spur gears were designed with a module of 1.5~mm, with a worm pitch diameter of 23.5~mm and a spur gear pitch circle diameter of 60~mm. Two SUS304 stainless steel springs of free length 50~mm are installed in parallel to the worm gear. Other components, such as the support structures, actuation pulley, and output link, are also fabricated from PLA. The Bowden cable unit used to drive the joint consists of a Polytetrafluoroethylene (PTFE) tube-based sheath with a polyester cable running through it. A laser displacement sensor (IL100, Keyence) was installed to measure the worm gear's linear displacement, which serves as feedback for force-based joint control.

While the proposed mechanism shares functional similarity with a series elastic actuator (SEA) with variable spring preload, its mechanical configuration is distinct. In conventional SEAs, the spring is placed directly in series between the angle control motor and the output, allowing transmission of motor torque through the elastic element. In contrast, WAVE incorporates a non-backdrivable worm gear that mechanically decouples the angle control motor from external forces. Although functionally similar to a variable-preload SEA, WAVE introduces a novel approach to stiffness modulation by distinctly decoupling the angle control motor from external forces.

\subsection{Stiffness model}
\begin{figure}
\centering
\includegraphics[width=0.9\linewidth]{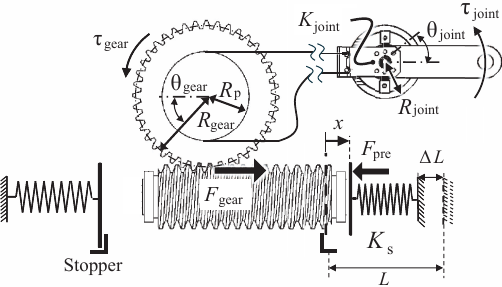}
\caption{System model of WAVE.}
\label{fig:system}
\end{figure}

This subsection introduces WAVE's mechanical model that explains how the stiffness control motor adjusts the spring precompression length ($\Delta L$) to set the target joint stiffness ($K_{joint}$), following the model \cite{jeong2021dual}. Fig.~\ref{fig:system} illustrates the forces, displacements, and torques within the mechanism.

Assuming two identical, ideal springs with a free length $L$ are compressed by a displacement $\Delta L$, the precompression force applied by the stiffness control motor can be expressed as:
\begin{equation}
F_{\mathrm{pre}} = K_{\mathrm{s}} \times \Delta L,
\label{eqn1}
\end{equation}
where $K_{\mathrm{s}}$ is the equivalent linear stiffness of the springs. The stiffness control motor modulates this force by adjusting the axial displacement $\Delta L$, as depicted in Fig.~\ref{fig:precompression}. When the joint is subjected to an external torque $\tau_{\mathrm{joint}}$, the corresponding external force transmitted through the cable is given by:
\begin{equation}
F_{\mathrm{gear}} = \frac{\tau_\mathrm{gear}}{R_\mathrm{gear}} = \frac{\tau_{\mathrm{joint}}}{R_\mathrm{{gear}}}\mu G
\end{equation}
where $\tau_\mathrm{gear}$ is the torque applied to the spur gear, $R_\mathrm{gear}$ is the spur gear radius, and $\mu$ is the efficiency of the Bowden cable system, ranging from 0 to 1 ($\mu \approx 1$ indicates minimal friction losses). $G$ is the gear ratio defined as $R_{\text{p}} / R_{\text{joint}}$. $R_{\text{joint}}$ is the joint radius and $R_p$ is the pulley radius in WAVE mechanism connected to the joint via cables.

Deformation of the spring mechanism occurs only when the external force $F_{\mathrm{gear}}$ exceeds the precompression force $F_{\mathrm{pre}}$. Therefore, the system operates in two states depending on the magnitude of $F_{\mathrm{gear}}$ with spring displacement $x$ given by,
%
%

\begin{equation}
x =
\begin{cases}
0 & \text{if } F_{\text{gear}} \leq F_{\text{pre}} \\
\frac{F_{\text{gear}} - F_{\text{pre}}}{K_s} & \text{if } F_{\text{gear}} > F_{\text{pre}}
\end{cases}
\end{equation}

\begin{figure}
\centering
\includegraphics[width=8.5cm]{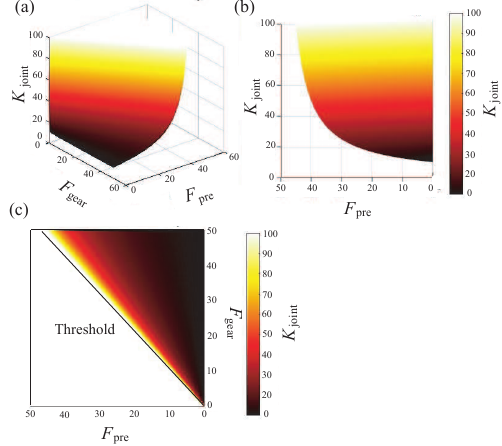}
\caption{Variation of joint stiffness with different spring precompression force and applied external force.}
\label{fig:model}
\end{figure}
%

The local linear stiffness $K$ of the system is given by:
\begin{equation}
K = \frac{F_{\text{gear}}}{x} =
\begin{cases}
\infty & \text{if } F_{\text{gear}} \leq F_{\text{pre}} \\
K_s \left(\frac{F_{\text{gear}}}{F_{\text{gear}} - F_{\text{pre}}} \right) & \text{if } F_{\text{gear}} > F_{\text{pre}}
\end{cases}
\label{eqn4}
\end{equation}

The resultant joint stiffness $K_\mathrm{joint}$ is defined as:

\begin{equation}
\begin{aligned}
K_{\mathrm{joint}} &= \displaystyle\frac{\tau_\mathrm{joint}}{\theta_\mathrm{joint}}
= \displaystyle\frac{\displaystyle\frac{\tau_\mathrm{gear}}{\mu G}}{\displaystyle\frac{x G}{R_{\mathrm{gear}}}}
= \displaystyle\frac{\displaystyle\frac{F_{\mathrm{gear}} R_{\mathrm{gear}}}{\mu G}}{\displaystyle\frac{x G}{R_{\mathrm{gear}}}} \\[5pt]
&= \displaystyle\frac{R_{\mathrm{gear}}^2}{\mu G^2} \cdot \displaystyle\frac{F_{\mathrm{gear}}}{x}
= \displaystyle\frac{R_{\mathrm{gear}}^2}{\mu G^2} \cdot K
\end{aligned}
\label{eqn5}
\end{equation}

Substituting Eq. (\ref{eqn1}) and Eq. (\ref{eqn4}) in Eq. (\ref{eqn5}),

\begin{equation}
\begin{aligned}
K_{\mathrm{joint}} =
\begin{cases}
\infty & \text{if } F_{\text{gear}} \leq K_s \Delta L \\[8pt]
\displaystyle \frac{R_{\mathrm{gear}}^2 K_s}{\mu G^2} \left( \frac{F_{\text{gear}}}{F_{\text{gear}} - K_s \Delta L} \right) & \text{if } F_{\text{gear}} > K_s \Delta L
\end{cases}
\end{aligned}
\label{eqn6}
\end{equation}

The equation reveals that joint stiffness $K_{joint}$ is a nonlinear function of spring precompression $\Delta L$ and external force $F_{gear}$, with the transition from rigid state to compliant state governed by the preload $F_{\text{pre}} = K_s \Delta L$. When the applied external force is less than the threshold $F_{\text{pre}}$, the springs remain uncompressed, and the joint is in a rigid state. As the applied force exceeds the threshold, the joint becomes compliant, and the stiffness asymptotically decreases and converges toward the spring constant. Thus, by adjusting the precompression length $\Delta L$ using the stiffness control motor, the joint stiffness can be effectively tuned--shifting the threshold, modifying the joint's response characteristics, and enabling continuous stiffness with respect to the applied external force.
%

Fig.~\ref{fig:model} presents simulated results based on this model, illustrating how $K_{\mathrm{joint}}$ varies with different values of $F_{\mathrm{pre}}$ and $F_{\mathrm{gear}}$. Notably, Fig.~\ref{fig:model}(b) confirms that when the applied force exceeds the preload threshold, the joint stiffness asymptotically converges to $K_{\mathrm{s}} R_{\mathrm{gear}}^2$, provided $G = \mu = 1$, in agreement with the theoretical model.

\section{EXPERIMENT}

In this section, we address the following key questions regarding the proposed WAVE mechanism:
\begin{enumerate}
    \item Does the joint successfully transition between the rigid and compliant states, and how does the precompression length $\Delta L$ affect joint stiffness?
    \item Can the worm-gear-based decoupling effectively protect the angle control motor from external force?
    \item How does the joint perform under common tasks, such as lifting and contact-rich operations?
\end{enumerate}


 \subsection{Validation of variable joint stiffness}



  \begin{figure}
  \centering
  \includegraphics[width=0.58\linewidth]{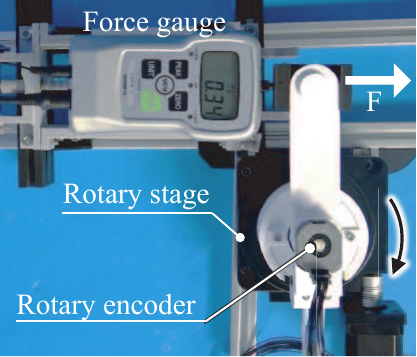}
  \caption{Experimental setup for joint performance evaluation.}
  \label{fig8}
  \end{figure}

   \begin{figure}
   \centering
   \includegraphics[width=0.88\linewidth]{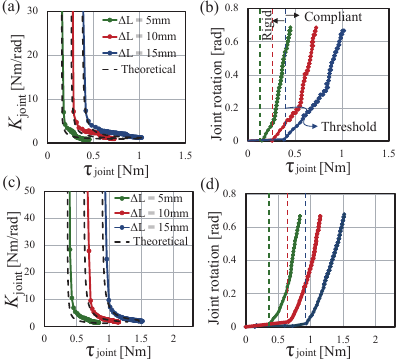}
   \caption{Stiffness of WAVE. (a) Joint stiffness vs torque and (b) Joint rotation vs torque characteristics for 0.8~N/mm spring. (c) Joint stiffness vs torque and (d) Joint rotation vs torque characteristics for 2.4~N/mm spring.}
   \label{fig10}
   \end{figure}

\begin{figure}
\centering
\includegraphics[width=0.75\linewidth]{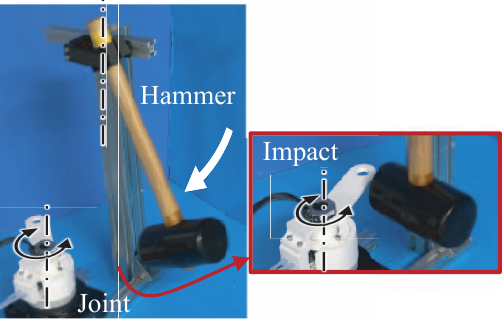}
\caption{Experimental setup for impact load application.}
\label{figham}
\end{figure}

This experiment aims to confirm how the WAVE's joint stiffness changes when external torque surpasses a threshold and how effectively precompression adjustments widen the stiffness range. The stiffness of WAVE was measured using the experimental setup shown in Fig.~\ref{fig8}, which consists of a rotary stage unit driven by a stepper motor (NEMA 17, 14HS17-0504S). The output link from the joint, with a moment arm of 100 mm, is attached to a force gauge (FGPX-5, Nidec SHIMPO) to measure the constraint force, and an absolute rotary encoder (CUI AMT203) is fixed to the joint to measure base rotation. When the motorized rotary base is actuated from 0$^\circ$ to 30$^\circ$, the overall joint undergoes rotation while the output link rotation is arrested by the force gauge to measure the external torque. The joint stiffness is determined by calculating the ratio of the applied external torque to the measured rotational angle.

Fig.~\ref{fig10}(a, b) shows the joint stiffness and rotational angle variation for different external torques applied to the joint with a spring stiffness of 0.8~N/mm, while Fig.~\ref{fig10}(c, d) shows the same for 2.4~N/mm. Fig.~\ref{fig10}(a) indicates that under threshold, joint stiffness remains relatively high. Once the external torque reaches the threshold, the stiffness follows a nonlinear, asymptotically decreasing trend, aligning with the simulated results shown by the black dotted line. The threshold depends on the spring precompression length, where higher precompression leads to higher thresholds, enhancing resistance to deformation. Theoretical thresholds ($K_{\text{s}}\cdot \Delta L \cdot R_{\text{gear}}$) were 0.12, 0.24, and 0.36~Nm for precompression lengths of 5, 10, and 15~mm, respectively. Experimentally, the joint operated in the range of 0.18--0.38~Nm, slightly higher due to cable friction and other losses. To evaluate the time response, we measured the time taken by the stiffness control motor to change precompression from 5 to 15~mm, which took approximately 2~s. As the stiffness control motor only actuates during precompression and stays idle otherwise, energy consumption is minimal. The peak current during modulation was 110 mA, with negligible idle current.


Fig.~\ref{fig10}(b) shows the trend of external torque variation with respect to joint rotation. Initially (pre-threshold), the joint behaves as a rigid body due to spring preload, resulting in negligible angular displacement. Once the applied torque exceeds the precompression threshold, the joint enters the compliant state where the spring compresses and stiffness follows the nonlinear profile described in Eq. (\ref{eqn6}). Within this state, the torque--rotation relationship can appear locally linear over small intervals, resulting in two approximately linear segments, as stiffness asymptotically approaches the intrinsic spring stiffness. Increasing spring precompression increases the threshold torque at which the joint rotation begins, thereby extending the range of rigid behavior. The experimental results showed not only an extension of this rigid state, but also a change in the relationship between joint torque and joint angle. Specifically, by adjusting the preload, the amount of joint deflection relative to the applied torque varied, demonstrating the mechanism's ability to exhibit variable stiffness.

Fig.~\ref{fig10}(c, d) presents the results for a spring stiffness of 2.4~N/mm, showing a similar trend to the lower stiffness case but with an increased threshold range of 0.4--0.9~Nm across precompression lengths from 5 to 15~mm. These results confirm the dual-mode behavior of the joint: A rigid state until the external torque is less than the threshold and a compliant state when the external force exceeds the threshold. Across both tested spring stiffness values (0.8 and 2.4~N/mm), the joint demonstrated thresholds within 0.18--0.9~Nm for precompression lengths of 5--15~mm. In the current configuration, the spring is precompressed up to 15~mm to realize high-stiffness behavior, while the maximum allowable compression is 30~mm. This leaves an additional 15~mm of available compression, beyond which the spring reaches its physical limit as the coils become fully compacted and can no longer deform.

  \begin{figure}
  \centering
  \includegraphics[width=0.92\linewidth]{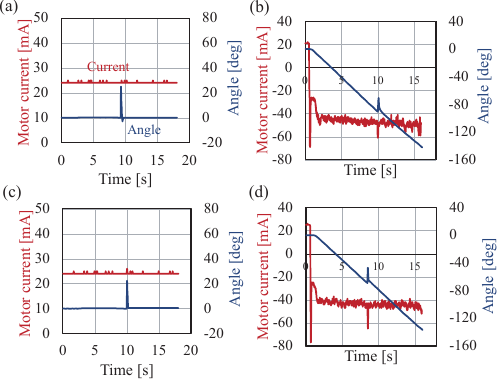}
  \caption{Joint response under impact load. (a) Variation of the angle control motor current and joint angle under impact when joint is at rest, and (b) During rotation (low-stiffness mode). (c) Variation of the angle control motor current and joint angle under impact when joint is at rest, and (d) During rotation (high-stiffness mode).}
  \label{fignew1}
  \end{figure}

  \begin{figure}
  \centering
  \includegraphics[width=0.91\linewidth]{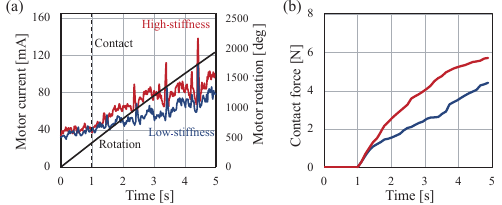}
  \caption{Joint response during contact. (a) Angle control motor current profile, and (b) Joint contact force under low- and high-stiffness modes.}
  \label{fig14}
  \end{figure}

\subsection{Impact absorption and loading tests}

Impact absorption and loading tests were conducted to evaluate how effectively the joint absorbs sudden impacts and manages contact forces under both low- and high-stiffness configurations. These configurations use springs with a stiffness of 2.4~N/mm, with precompression lengths of 5~mm and 15~mm corresponding to the low- and high-stiffness modes, respectively. To assess impact response, the joint was subjected to a hammer-induced collision using the experimental setup illustrated in Fig.~\ref{figham}. The hammer was manually pulled back to an angle of 45$^\circ$ and released at the moment when the joint's output link aligned with the hammer's base. The force from the hammer impact was estimated to be approximately 20~N, as measured by the force gauge, which exerted a torque of 2~Nm on the joint.

Fig.~\ref{fignew1} shows the variation in angle control motor current relative to joint actuation under impact load. As shown in Fig.~\ref{fignew1}(a, c), when the joint experiences an impact while at rest, the joint angle spikes due to spring compression and then returns to its original position from spring tension. During impact, the load is transferred to the springs via sliding worm gear, ensuring the angle control motor remains protected in both stiffness cases, as verified by Fig.~\ref{fignew1}(a, c). This confirms the decoupling of external forces. When the joint with low-stiffness is rotating and encounters an impact, the rotation angle and motor current momentarily spike due to external resistance. However, the joint quickly becomes compliant and transfers the majority of the load to the springs, safeguarding the angle control motor both in low- and high-stiffness modes showcasing effective impact absorption as shown in Fig.~\ref{fignew1}(c, d).
To quantify this, motor current spikes and joint displacement during impact were analyzed for both stiffness modes. In both cases, the non-backdrivable worm gear decouples the angle control motor from external disturbances and the axial sliding of worm gear transferred the shock to the spring, with little to no current spikes on the angle control motor--thereby validating WAVE's ability to decouple impacts and reduce motor stress.


An experiment was conducted to analyze the joint's behavior during a contact-based task using the same setup shown in Fig.~\ref{fig8}, where a barrier was fixed to a force gauge to measure contact force as the output link impacted the barrier under active joint actuation. Upon contact, the spur gear's rotation halted, causing the worm gear to slide passively under continued joint drive. A sharp rise in contact force was expected at the moment of contact, since worm gear displacement occurs only after overcoming the spring's precompression.
Fig.~\ref{fig14}(a) shows the variation in the angle control motor current upon contact, where the motor was rotated five times after initial contact, and measurements were recorded. With low joint stiffness, most of the motor's torque was used to compress the springs, with only a small portion transferred to the output link and some dissipated due to internal worm gear friction. This resulted in a contact force of approximately 4.2~N and a motor current spike of 45~mA. In the high-stiffness case, resistive contact forces increased sharply, and the motor current peaked at 85~mA. The higher spring precompression directed more torque to the output link, producing a higher contact force of 6~N with minimal spring compression. These results indicate that the angle control motor operates more safely in low-stiffness mode when rotating against rigid objects such as walls. Thus, upon contact in both low- and high-stiffness modes, much of the angle control motor's torque was directed toward spring compression and cable friction, thereby maintaining safe contact force.


\subsection{Position tracking and control performance}

\begin{figure}
    \centering
    \includegraphics[width=0.9\linewidth]{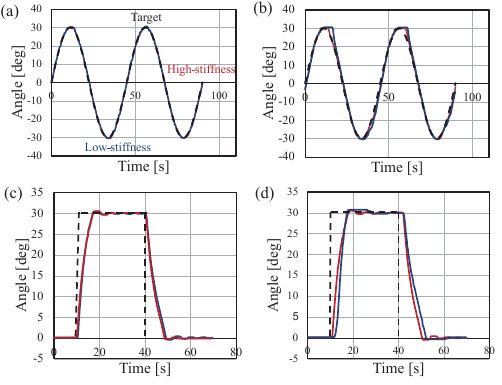}
    \caption{Position control of the joint. (a) Sinusoidal response under lateral loading. (b) Sinusoidal response under vertical loading. (c) Step response under lateral loading. (d) Step response under vertical loading.}
    \label{fig12}
\end{figure}

 \begin{figure}
    \centering
    \includegraphics[width=0.9\linewidth]{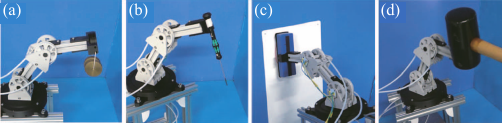}
    \caption{ Applications of manipulator with WAVE. (a) Payload handling (b) Tool manipulation (c) Wiping and (d) Shock absorption.}
    \label{fig17}
\end{figure}

%

We aim to quantify the joint's position tracking accuracy and control stability under varying load conditions and stiffness settings. The system was operated under low-speed to (i) minimize cable backlash effects, (ii) ensure accurate stiffness estimation, and (iii) allow visible tracking of spring compliance. Additionally, the high reduction ratio of the worm gear (40: 1) further reduces the output speed. The joint was driven in a closed-loop sinusoidal motion with a 30$^\circ$ amplitude and a 45~s period. Fig.~\ref{fig12}(a, b) shows results under 0.5~kg lateral and vertical loading, applying joint torque within the threshold range of low- and high-stiffness. The joint maintained higher accuracy under lateral loading, with a maximum error of 5.6$^\circ$ in the low-stiffness case. A 30$^\circ$ step input test under the same loading conditions was also performed, as shown in Fig.~\ref{fig12}(c, d). The oscillatory response lasted about 12~s in low-stiffness and 5~s in high-stiffness, reflecting the joint's compliant nature. Positional accuracy remained satisfactory, with a maximum error of 1.5$^\circ$, closely following the desired trajectory under lateral load. The improved performance under lateral loading is attributed to reduced spring compression, whereas vertical loads induce greater deformation. Overall, tracking accuracy was acceptable in both stiffness modes, with lateral loads resulting in slightly better control performance.


\subsection{Practical demonstrations using a manipulator with WAVE}
%

We developed a 3-DOF robotic manipulator using the WAVE actuator and conducted experiments to evaluate its feasibility in manipulation tasks, as shown in Fig.~\ref{fig17}.
The images illustrate various scenarios where the joint's adaptable stiffness improves functionality and safety.
(a) Payload handling: The arm handles a 300~g object with all joints in high-stiffness mode, demonstrating stable lifting and placement.
(b) Tool manipulation: The arm is equipped with a tool, showcasing its ability to perform tool-based tasks.
(c) Wiping: The manipulator wipes a whiteboard using joints with low-stiffness, enabling safe performance of contact-rich tasks.
(d) Shock absorption: The arm is subjected to an external impact using a hammer, demonstrating the joint's ability to absorb shocks and protect internal components. These applications demonstrate that manipulators equipped with WAVE can leverage variable stiffness control to handle tasks efficiently, absorb impacts, and enhance overall safety.

\section{LIMITATION}

While experimental results demonstrate WAVE's advantages, it also has limitations due to the use of friction-based transmission.
Under high loads, the transmission efficiency of the worm gear decreases, potentially reducing both speed and torque compared to direct joint actuation. Additionally, sudden impact forces may cause momentary locking of the worm gear at high-stiffness, leading to stalling of the angle control motor. Although the maximum load on the motor is limited to its stall torque, reducing the risk of damage, addressing friction-related losses remains a challenge for future improvements. Future work will focus on mitigating WAVE's friction-related losses and worm gear locking under load by utilizing a switchable coefficient of friction through vibration [7], [22] or a volatile lubricant [23].

\section{CONCLUSION}

This paper proposed WAVE, which uses a worm gear to enable stiffness modulation and protect the actuators by decoupling them from external forces. Experimental results confirmed that WAVE effectively varies stiffness, decouples the angle control motor from external forces, and absorbs impacts with low-stiffness while supporting heavy payloads with high-stiffness. Thus, WAVE allows extended operation in contact-intensive tasks, making it suitable for robust robotic applications in challenging environments.
Future improvements using high-power motors and metal components can improve WAVE's strength and torque capabilities.


\begin{thebibliography}{99}

\bibitem{wolf2015variable}
S. Wolf, G. Grioli, O. Eiberger, W. Friedl, M. Grebenstein, H. H\"oppner, E. Burdet, D. G. Caldwell, R. Carloni, M. G. Catalano, N. G. Tsagarakis, A. Bicchi, and A. Albu-Sch\"affer, ``Variable stiffness actuators: Review on design and components,'' \emph{IEEE/ASME Trans. Mechatronics}, vol. 21, no. 5, pp. 2418--2430, 2015.

\bibitem{vanderborght2013variable}
B. Vanderborght, A. Albu-Sch\"affer, A. Bicchi, E. Burdet, D. G. Caldwell, R. Carloni, M. Catalano, O. Eiberger, W. Friedl, G. Ganesh, M. Garabini, M. Grebenstein, H. H\"oppner, J. Morimoto, T. Wimb\"ock, S. Wolf, and N. G. Tsagarakis, ``Variable impedance actuators: A review,'' \emph{Robotics and Autonomous Systems}, vol.61, no.12, pp.1601--1614, 2013.

\bibitem{albu2007dlr}
A. Albu-Sch\"affer, S. Haddadin, C. Ott, A. Stemmer, T. Wimb\"ock, and G. Hirzinger, ``The DLR lightweight robot: design and control concepts for robots in human environments,'' \emph{Industrial Robot: An International Journal}, vol. 34, no. 5, pp. 376--385, 2007.

\bibitem{schiavi2008vsa}
R. Schiavi, G. Grioli, S. Sen, and A. Bicchi, ``VSA-II: A novel prototype of variable stiffness actuator for safe and performing robots interacting with humans,'' in \emph{Proc. IEEE Int. Conf. Robotics and Automation}, 2008, pp. 2171--2176.

\bibitem{catalano2011vsa}
M. G. Catalano, G. Grioli, M. Garabini, F. Bonomo, M. Mancini, N. G. Tsagarakis, and A. Bicchi, ``VSA-Cubebot: A modular variable stiffness platform for multiple degrees of freedom robots,'' in \emph{Proc. IEEE Int. Conf. Robotics and Automation}, 2011, pp. 5090--5095.

\bibitem{li2024novel}
Z. Li, P. Xu, and B. Li, ``A novel variable stiffness actuator based on cable-pulley-driven mechanisms for robotics,'' \emph{IEEE/ASME Trans. Mechatronics}, 2024.

\bibitem{takayama2019worm}
T. Takayama and N. Hisamatsu, ``Worm gear mechanism with switchable backdrivability,'' \emph{Robomech Journal}, vol. 6, pp. 1--10, 2019.

\bibitem{eiberger2010qa}
O. Eiberger, S. Haddadin, M. Weis, A. Albu-Sch\"affer, and G. Hirzinger, ``On joint design with intrinsic variable compliance: Derivation of the DLR QA-joint,'' in \emph{Proc. IEEE Int. Conf. Robotics and Automation}, 2010, pp. 1687--1694.

\bibitem{vo2020compact}
C. P. Vo, V. D. Phan, T. H. Nguyen, and K. K. Ahn, ``A compact adjustable stiffness rotary actuator based on linear springs: working principle, design, and experimental verification,'' \emph{Actuators}, vol. 9, no. 4, p. 141, 2020.

\bibitem{visser2011energy}
L. C. Visser, R. Carloni, and S. Stramigioli, ``Energy-efficient variable stiffness actuators,'' \emph{IEEE Trans. Robotics}, vol.27, no.5, pp.865--875, 2011.

\bibitem{tsagarakis2011compact}
N. G. Tsagarakis, I. Sardellitti, and D.G.Caldwell, ``A new variable stiffness actuator (CompAct-VSA): Design and modelling,'' in \emph{Proc. IEEE/RSJ Int. Conf. Intelligent Robots and Systems},2011,pp.378--383.

\bibitem{sariyildiz2023design}
E. Sariyildiz, R. Mutlu, J. Roberts, C. H. Kuo, and B. U\u{g}urlu, ``Design and control of a novel variable stiffness series elastic actuator,'' \emph{IEEE/ASME Trans. Mechatronics}, vol. 28, no. 3, pp. 1534--1545, 2023.

\bibitem{english1999mechanics}
C. English and D. Russell, ``Mechanics and stiffness limitations of a variable stiffness actuator for use in prosthetic limbs,'' \emph{Mechanism and Machine Theory}, vol. 34, no. 1, pp. 7--25, 1999.

\bibitem{jeong2015dual}
H. Jeong, J. Cheong, and S. Kwon, ``Dual-mode variable stiffness actuator using two-stage worm gear transmission for safe robotic manipulators,'' \emph{Int. J. Precision Eng. Manufacturing}, vol. 16, pp. 1761--1769, 2015.

\bibitem{tsagarakis2013asymmetric}
N. G. Tsagarakis, S. Morfey, H. Dallali, G. A. Medrano-Cerda, and D. G. Caldwell, ``An asymmetric compliant antagonistic joint design for high performance mobility,'' in \emph{Proc. IEEE/RSJ Int. Conf. Intelligent Robots and Systems}, 2013, pp. 5512--5517.

\bibitem{moore2021quadratic}
R. Moore and J. M. Schimmels, ``Design of a quadratic, antagonistic, cable-driven, variable stiffness actuator,'' \emph{J. Mechanisms and Robotics}, vol. 13, no. 3, p. 031001, 2021.

\bibitem{tonietti2005design}
G. Tonietti, R. Schiavi, and A. Bicchi, ``Design and control of a variable stiffness actuator for safe and fast physical human/robot interaction,'' in \emph{Proc. IEEE Int. Conf. Robotics and Automation}, 2005, pp. 526--531.

\bibitem{zhou2015cable}
X. Zhou, S. K. Jun, and V. Krovi, ``A cable based active variable stiffness module with decoupled tension,'' \emph{ASME J. Mechanisms and Robotics}, vol. 7, no. 1, p. 011005, 2015.

\bibitem{wockenfuss2022design}
W. R. Wockenfu{\ss}, V. Brandt, L. Weisheit, and W.-G. Drossel, ``Design, modeling and validation of a tendon-driven soft continuum robot for planar motion based on variable stiffness structures,'' \emph{IEEE Robotics and Automation Letters}, vol. 7, no. 2, pp. 3985--3991, 2022.

\bibitem{ranzani2013modular}
T. Ranzani, M. Cianchetti, G. Gerboni, I. De Falco, G. Petroni, and A. Menciassi, ``A modular soft manipulator with variable stiffness,'' in \emph{Joint Workshop on New Technologies for Computer/Robot Assisted Surgery}, 2013, pp. 11--13.

\bibitem{jeong2021dual}
H. Jeong, J. Cheong, and S. Kwon, ``Dual-mode variable stiffness actuator using two-stage worm gear transmission for safe robotic manipulators,'' \emph{International Journal of Precision Engineering and Manufacturing}, vol. 16, pp. 1761--1769, Jul. 2015.

\bibitem{hu2023influence}
S. Hu, H. She, G. Yang, C. Zang, and C. Li, ``The influence of interface roughness on the vibration reduction characteristics of an under-platform damper,'' \emph{Applied Sciences}, vol.13, no.4, p.2128, 2023.

\bibitem{nishimura2022soft}
T. Nishimura, K. Shimizu, S. Nojiri, K. Tadakuma, Y. Suzuki, T. Tsuji, and T. Watanabe, ``Soft robotic hand with finger-bending/friction-reduction switching mechanism through 1-degree-of-freedom flow control,'' \emph{IEEE Robotics and Automation Letters}, vol. 7, no. 2, pp. 5695--5702, 2022.

\end{thebibliography}
\end{document}